# Glioma Grade Prediction Using Wavelet Scattering-Based Radiomics

**Qijian Chen [1], Lihui Wang[1\*], Li Wang [1], Zeyu Deng [1], Jian Zhang [1], Yuemin Zhu[2]**

[1]Key Laboratory of Intelligent Medical Image Analysis and Precise Diagnosis of Guizhou Province, College of Computer Science and Technology, Guizhou University, Guiyang, 550025.

[2]CNRS UMR5220; Inserm U1206; INSA Lyon; Université de Lyon, France.

corresponding author: Lihui Wang, (email: lhwang2@gzu.edu.cn, wlh1984@gmail.com)

This work is funded partially by the National Nature Science Foundations of China (Grant No. 61661010), the Nature Science Foundation of Guizhou province (Qiankehe J No.20152044), the Program PHC-Cai Yuanpei 2018 (N° 41400TC) and the Guizhou Science and Technology Plan Project (Qiankehe [2018]5301).

**ABSTRACT** Glioma grading before surgery is very critical for the prognosis prediction and treatment plan making. We present a novel wavelet scattering-based radiomic method to predict noninvasively and accurately the glioma grades. The method consists of wavelet scattering feature extraction, dimensionality reduction, and glioma grade prediction. The dimensionality reduction was achieved using partial least squares (PLS) regression and the glioma grade prediction using support vector machine (SVM), logistic regression (LR) and random forest (RF). The prediction obtained on multimodal magnetic resonance images of 285 patients with well-labeled intratumoral and peritumoral regions showed that the area under the receiver operating characteristic curve (AUC) of glioma grade prediction was increased up to 0.99 when considering both intratumoral and peritumoral features in multimodal images, which represents an increase of about 13% compared to traditional radiomics. In addition, the features extracted from peritumoral regions further increase the accuracy of glioma grading.

## I. INTRODUCTION

Gliomas are the most common primary malignant tumors of the central nervous system (CNS), which have high incidence, recurrence, mobility and mortality rate. How to treat the gliomas effectively is still a challenge. Generally, gliomas can be classified into low-grade gliomas (LGG) and high-grade gliomas (HGG) [1]. Different grades correspond to different surgical plans and radiotherapy or chemotherapy strategies. Therefore, accurate grading prediction plays an important role in the treatment-decision making, personalized patient management, and prognostic evaluation. Currently, biopsy or histopathological assessment after surgery is considered the gold standard for glioma grading [2]. However, such grading means is invasive, time-consuming, painful and useless for those patients not suitable for the surgery. Therefore, developing a noninvasive strategy for precisely grading gliomas is essential.

Medical imaging, especially magnetic resonance imaging (MRI), is a promising noninvasive tool for characterizing gliomas. It was shown that contrast-enhanced T1-weighted imaging (T1-CE), diffusion-weighted imaging (DWI) and arterial spin labeling (ASL) imaging have great potential in gliomas grading by noninvasively exploring the heterogeneity of tumors from a microscopic view [3],[4],[5], for example, with apparent diffusion coefficient (ADC) to reflect the cell density of gliomas, with perfusion coefficients and perfusion fraction to reveal the variation of vascular components in gliomas. Ryu et al. proved that texture analysis of ADC maps in DWI is useful for evaluating glioma grade [6]. Osamu et al. demonstrated that intravoxel incoherent motion (IVIM) imaging is helpful for differentiating HGG gliomas from LGG gliomas [7]. However, extracting information from a single modality or simple comparison among different modalities is not enough for accurate grading analysis.

Recently, radiomics becomes an emerging noninvasive method to quantitatively characterize medical images by extracting high-throughput image features from multiple imaging modalities, including shapes, textures, wavelet features, etc. [8],[9]. It has been successfully used for phenotypic analysis and prognosis prediction of several tumors [10],[11],[12]. However, there are still few works on using radiomics to predict glioma grades. Brynolfsson et al. [13] demonstrated that the gray level co-occurrence matrix (GLCM)-based texture features are useful for glioma grading and prognosis prediction. Following that, Cho et al. showed that the combination of histogram and GLCM-based texture features performs better in distinguishing low-grade and high-grade gliomas [14]. To promote prediction accuracy, the researchers investigated several feature selection methods and classification models to get a higher prediction accuracy [15],[16],[17].

Due to the excellent performance of deep convolutional network (CNN) in image classification, some researchers also used deep learning models to predict glioma grade. For instance, Ge et al proposed a saliency-aware strategy to enhance tumor regions of MRIs and used a novel feature fusion scheme for classifying high- and low-grade gliomas with accuracy of 89.47% [40]. They further designed a multi-stream deep CNN architecture to improve the performance of glioma grading with T1, T2 and FLAIR images, the accuracy was up to 90.87% [41]. To deal with the influence of the sample insufficiency, Ali et al attempted to use generative adversarial networks (GANs) for data augmentation and then used convolutional autoencoder (CAE) to classify high- and low-grade gliomas, the accuracy was increased to 92.04%.[42]. Although deep neural networks show excellent performance in various tasks, interpretability is still a problem to deal with, especially for clinical applications [43]. In addition, the features learned by CNN lack the invariance [44], especially not robust to the noise, therefore the performance may be not stable.

Image features are the most important factors that influence the prediction ability of radiomic methods. Traditional features used in radiomics, such as textural and wavelet features, are very sensitive to noise, transformations (translation and rotation) and small deformations. For instance, if a small region containing the same tumor is located at different positions in the image, traditional features like textural and wavelet features extracted from the region will be very different, depending on its position, which consequently influences the prediction accuracy of radiomics. Therefore, how to extract local transformation-invariant and noise-insensitive features to increase prediction ability of radiomics is still a challenge.

In view of the interest of wavelet scattering for representing invariant image features [18], [28], we propose to use wavelet scattering radiomic features instead of traditional radiomic features to noninvasively predict glioma grading. Wavelet scattering features are robust both to noise due to the involved image high-frequency averaging and to local transformations owing to the use of the mean value of scattering feature maps. Furthermore, most radiomics-based glioma grading studies focused on intratumoral regions, such as necrotic, non-enhancing solid and enhancement core of the tumor. The surrounding environment of the tumor remains unexplored; it may however provide unique information for glioma grading. All that led us to propose a novel wavelet scattering-based radiomic method. The idea is to extract wavelet scattering features from different tumor-related regions in the images coming from different imaging modalities, use partial least squares (PLS) regression to reduce the number of the extracted features, and predict glioma grades by means of classifiers such as support vector machine (SVM), logistic regression (LR) and random forest (RF).

## II. MATERIALS AND METHODS

### A. DATA DESCRIPTION

The data used in this work was downloaded from the MICCAI website 2017 targeting for the glioma segmentation challenge [19], [20], [21], which is classified into 75 patients with LGG (astrocytoma or oligo-astrocytoma) and 210 patients with HGG (anaplastic astrocytoma and glioblastoma multiforme tumors) based on histological diagnosis. All the patients were examined with axial T1-weighted, T1-Gd enhanced and T2-weighted images. To overcome the influence of the motions of patients, the skull stripping was performed firstly, followed by image registration to make sure that the multi-contrast images are strictly matched for the same patient [19], [20], [21]. The spatial resolution of the registered images is 1 mm ×1 mm ×1 mm [22]. The regions of interest (ROIs) were drawn manually by experienced radiologists, including edema, non-enhancing solid core, necrotic core, and enhancing core. In the present work, to analyze the influence of different tumor regions on the prediction accuracy of glioma grade, for simplicity, we considered that the necrotic core, non-enhancing solid core and enhancing core consist of intratumoral region, and that the edema that excludes intratumoral part constitutes the peritumoral region, as illustrated in Fig. 1.

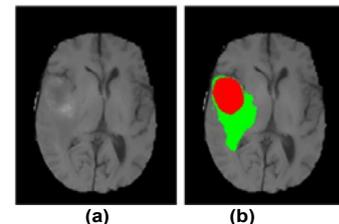

**FIGURE 1.** Illustration of original image (a) and the corresponding ROIs (b). The red indicates the intratumoral region and the green the peritumoral region.

In addition to the images, the clinical properties of the subjects can be found in the Cancer Genome Atlas (TCGA) [23], as summarized in TABLE I.

TABLE I
CLINICAL FEATURES OF THE PATIENTS.

| Clinical Features | Value |
|---|---|
| No. of patients | 285 |
| Age (mean±std, years) | 60.33±12.08 |
| <30 | 3(1.0%) |
| 30-50 | 28(9.8%) |
| 50-70 | 98(34.4%) |
| 70-80 | 29(10.2%) |
| >80 | 5(1.8%) |
| NA | 122(42.8%) |
| Survival (mean±std) | 422.96±349.68 |
| <300 | 65(22.8%) |
| 300-1000 | 87(30.5%) |
| >1000 | 11(3.9%) |
| NA | 122(42.8%) |
| Tumor Grade | |
| High-grade (HGG) | 210(73.7%) |
| Low-grade (LGG) | 75(26.3%) |

### B. RADIOMICS BASED ON WAVELET SCATTERING

The proposed radiomic framework is composed of four main steps: ROI extraction, feature extraction, feature selection, and glioma grade prediction. Taking into account the insufficiency of traditional features, such as sensitivity to noise and local transformation, in the present work, we propose to use wavelet scattering features to replace traditional features to get more meaningful features for promoting prediction accuracy. The overall workflow of the prediction model is illustrated in Fig. 2.

The first-order, shape, texture, wavelet features and wavelet scattering (WS) features are respectively extracted from manual segmented ROIs. The extracted traditional features and wavelet scattering features were selected using PLS method. After that, the selected features are fed into three classification models (SVM, LR and RF).

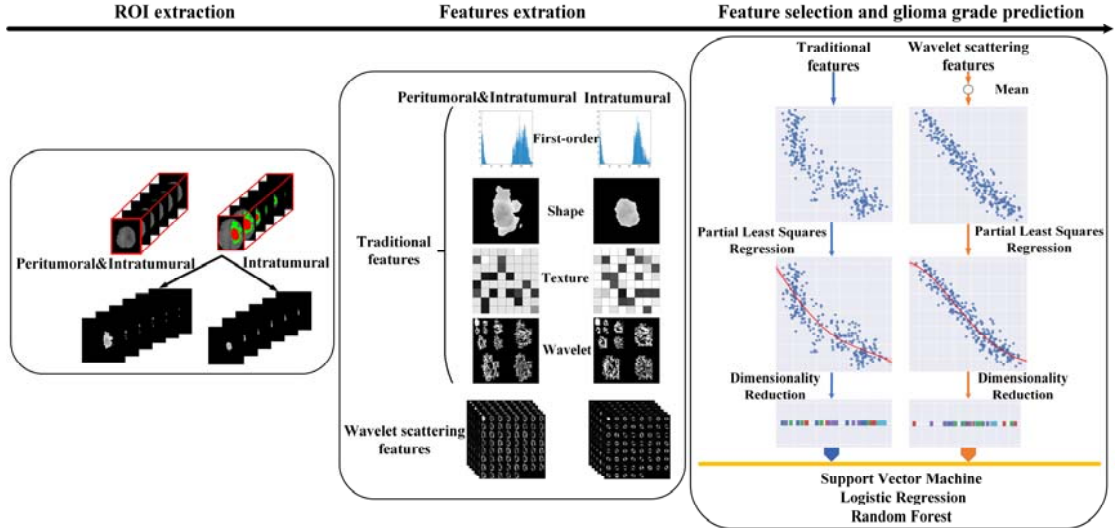

**FIGURE 2.** Overall workflow of glioma grade prediction based on traditional features and wavelet scattering features.

Wavelet scattering is developed based on wavelet transform that is devoted to analyzing images from a multiscale point of view and extracts image features by convolving variants of filters with image $x$. That is

$$Wx = \begin{bmatrix} x * \phi_J \\ x * \psi_{j,r} \end{bmatrix},$$ (1)

where $\psi_{j,r}(u) = 2^{-2j}\psi(2^{-j}r^{-1}u)$ represents the translation, scaling and rotating of wavelet function with the scaling factor $j$ satisfying that $1 \le 2^j \le 2^J$ ($J$ is the maximum scaling index) and $r = 2\pi l / L$ denoting the rotation angle of wavelet function ($L$ is the maximum number of rotations, $l = 0,1...L$), $*$ designates convolution operator. Various wavelet functions can be used to extract the high-frequency information of images. Generally, the scaling function $\phi_J$ is composed of a series of Gaussian functions and is dedicated to expressing the low-frequency information of images. It is formulated as:

$$\phi_J(u) = 2^{-2J}\phi(2^{-J}u),$$ (2)

with $\phi$ being expressed by Gaussian function

$$\phi(u) = e^{-u^2/2\sigma^2}.$$ (3)

Although the wavelet function is able to restore image details, it is only translation-invariant at the current scale of $2^j$ due to its localization properties [18]. In order to extend this translation-invariant property to the biggest scale of $x$ and keep simultaneously the stability for deformation, the averaging operation at the scale of $2^J$ is performed by convolving the high-frequency coefficient and the low-frequency filter, namely $x * \psi_{j,r} * \phi_J$. However, the result of such convolution is zero because the wavelet function $\psi_{j,r}$ and the scaling function $\phi_J$ are orthogonal. This implies that no information will be generated by averaging directly the high-frequency coefficients. To deal with this issue, a nonlinear operation, namely the modulus of high-frequency coefficients is achieved before the averaging [24]. Then, the translation-invariant features can be obtained by

$$\left| x * \psi_{j,r} \right| * \phi_J.$$ (4)

From (4) we can see that the high-frequency information is lost after the low-pass filtering. To recover the high-frequency information, wavelet decomposition at larger scale (must be smaller than the biggest scale $J$) is performed on the modulus of the current high-frequency coefficient, which can be formulated as

$$\left| x * \psi_{j,r} \right| * \psi_{j+1,r}.$$ (5)

However, as mentioned above, such high-frequency information at the current scale is not translation-invariant. To keep the translation-invariant coefficients, a modulo operation followed by a low-pass filter should be performed again:

$$\left| \left| x * \psi_{j,r} \right| * \psi_{j+1,r} \right| * \phi_J.$$ (6)

From (5) and (6), it can be observed that the translation-invariants are obtained by applying wavelet transform on the modulus of high-frequency coefficients followed by an averaging operation. This process is called wavelet scattering transform, which can be expressed as:

$$\widetilde{W}x = \begin{bmatrix} S_m \\ U_m \end{bmatrix}. \tag{7}$$

Define the wavelet scattering propagator $U_m$ as:

$$
\begin{aligned}
&U_m(p\{\lambda_{j_0,r}, \ldots, \lambda_{j_m,r}\}) \\
&= \left| \ldots \left| \left| x * \psi_{j_0,r} \right| * \psi_{j_1,r} \right| \ldots * \psi_{j_m,r} \right| \\
&= \left| U_{m-1}(p\{\lambda_{j_0,r}, \ldots, \lambda_{j_{m-1},r}\}) * \psi_{j_m,r} \right| \\
&\text{with} \quad U_0 = \left| x * \psi_{j_0,r} \right|, \ m = 1,2\ldots M
\end{aligned}
\tag{8}
$$

where $\lambda_{j_m,r} = (j,m,r)$ records the scale information $j_m$ and rotation direction $r$ for the scattering level $m$, $p\{\lambda_{j_0,r}, \ldots, \lambda_{j_m,r}\}$ represents the scattering propagation path for the scattering level $j_m$ at scale $j_m$ along the direction $r$, and $M$ is the maximum scattering layer. Accordingly, the scattering propagator matrix $U$ can be written as:

$$
U = \begin{bmatrix} U_0 = \left| x * \psi_{j0,r} \right| \\ U_1 = \left| U_0 * \psi_{j_1,r} \right| \\ \vdots \\ U_m = \left| U_{m-1} * \psi_{j_m,r} \right| \end{bmatrix}. \tag{9}
$$

The corresponding wavelet scattering coefficient $S_m(p\{\lambda_{j_1,r}, \ldots, \lambda_{j_m,r}\})$ is:

$$S_m(p\{\lambda_{j_1,r}, \ldots, \lambda_{j_m,r}\}) = U_{m-1}(p\{\lambda_{j_1,r}, \ldots, \lambda_{j_m,r}\}) * \phi_J \tag{10}$$

The detailed expression for $S_m(p\{\lambda_{j_1,r}, \ldots, \lambda_{j_m,r}\})$ is given by:

$$
\begin{aligned}
S &= \begin{bmatrix} S_0 = x * \phi_J \\ S_1 = U_0 * \phi_J \\ \vdots \\ S_m = U_{m-1} * \phi_J \end{bmatrix} \\
&= \begin{bmatrix} S_0 = x * \phi_J \\ S_1 = \left| x * \psi_{j_0,r} \right| * \phi_J \\ \vdots \\ S_m = \left| \ldots \left| \left| x * \psi_{j_0,r} \right| * \psi_{j_1,r} \right| \ldots * \psi_{j_{m-1},r} \right| * \phi_J \end{bmatrix}
\end{aligned}
\tag{11}
$$

During the wavelet scattering decomposition, the scale of wavelet scattering should satisfy $j_1 < j_2 < \cdots < j_m < J$.

According to the above wavelet scattering principle, we extracted invariant features using the following parameters: the number of wavelet scattering levels is $m=2$, the wavelet decomposition scale is $J=2$, and the scattering direction at each scale is $L=4$ which results in $r=[0, \pi/2, \pi, 3\pi/2]$. With such parameter setting, the scheme of the second-order scattering network was illustrated in Fig. 3, in which, $j$ indicates the wavelet transform level, $m$ the scattering order, and the arrows the propagating paths. Along each path, there is a scattering propagator expressed as $U$. The blue arrows represent the zero-order scattering propagating paths, along which the low-frequency image is decomposed continuously to the maximum scale. The red arrows express the first-order scattering propagating paths, along which the propagators are obtained by decomposing once the modulus of high-frequency image. The green arrows denote the second-order scattering propagating paths; the corresponding propagators decompose twice the modulus of high frequency image. According to the principle of wavelet scattering, the $m^{th}$-order scattering features are obtained by convolving the $m$-$1^{th}$ order scattering propagators with scaling function $\phi_J$, marked as the black arrows in the bottom of Fig. 3.

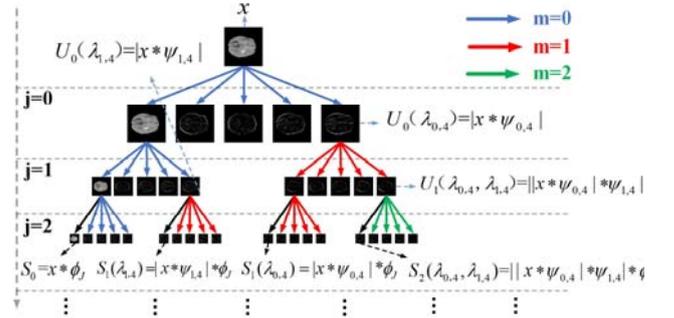

**FIGURE 3.** Scheme of the second-order wavelet scattering network.

The final outputs constitute the scattering representation of image $x$. If $m=2$, we need to calculate the invariant features $S_0$, $S_1$ and $S_2$. According to (9) and (11) as well as the constraint $j_0 < j_1 < \cdots < j_m < J$, the propagation operator $U_0$ in four directions can be formulated as:

$$
\begin{aligned}
U_0 &= U_0(\lambda_{0,1}) + U_0(\lambda_{0,2}) + U_0(\lambda_{0,3}) + U_0(\lambda_{0,4}) \\
&\quad + U_0(\lambda_{1,1}) + U_0(\lambda_{1,2}) + U_0(\lambda_{1,3}) + U_0(\lambda_{1,4}) \\
&\text{with} \quad j_0 = 0,1
\end{aligned}
\tag{12}
$$

The resulting local invariant scattering feature $S_1$ is

$$
\begin{aligned}
S_1 &= U_0(\lambda_{0,1}) * \phi_J + U_0(\lambda_{0,2}) * \phi_J + U_0(\lambda_{0,3}) * \phi_J + U_0(\lambda_{0,4}) * \phi_J \\
&\quad + U_0(\lambda_{1,1}) * \phi_J + U_0(\lambda_{1,2}) * \phi_J + U_0(\lambda_{1,3}) * \phi_J + U_0(\lambda_{1,4}) * \phi_J
\end{aligned}
\tag{13}
$$

Similarly, the propagation operator for the scattering level 1 is

$$U_1 = \sum_{j_1=j_0+1}^{J} \sum_{j_0=0}^{j_1} \sum_{r=1}^{4} U_1(\lambda_{j_0,r}, \lambda_{j_1,r}), \tag{14}$$

where $U_1(\lambda_{j_0,r}, \lambda_{j_1,r}) = \left\| x * \psi_{j_0,r} \right| * \psi_{j_1,r} \right\|$. Due to the limit $j_0 < j_1 < J = 2$, we can derive that $j_0 = 0, j_1 = 1$. As a result,

$$U_1 = \sum_{r=1}^{4} U_1(\lambda_{0,r}, \lambda_{1,r}) = \sum_{r=1}^{4} \left\| x * \psi_{0,r} \right| * \psi_{1,r} \right|. \tag{15}$$

The corresponding wavelet scattering features are given by

$$S_2 = U_1 * \phi_J. \qquad (16)$$

Based on the above formulations, if we use two-level wavelet scattering, we can get a total of 25 scattering feature maps, and the numbers of $S_0, S_1$ and $S_2$ are 1, 8 and 16, respectively.

If $x$ is an image of $N^2$ pixels and the maximum number of rotations is $L$, the overall complexity to compute the wavelet scattering features is $O(L^2 N^2 \log(N))$.

### C. PREDICTION EVALUATION

The performance of the proposed wavelet scattering-based radiomics for glioma grade prediction was evaluated in terms of receiver operating characteristic (ROC) curve, area under curve (AUC), sensitivity, specificity and accuracy. ROC is obtained by plotting true positive rate (TPR) against false positive rate (FPR) at different thresholds in a classifier. AUC indicates the surface under the curve of ROC and specifies the classification accuracy. The bigger the AUC, the more accurate the classification. Sensitivity represents the correct classification rate of positive samples while specificity represents the correct classification rate of negative samples. These metrics allow reflecting the false positive and false negative errors of the prediction models. Since AUC is not sensitive to sample properties such as the unbalance of sample classes, it is often used to evaluate the performance of the classifier for unbalanced dataset. The proposed radiomics was also compared with traditional radiomics.

## III. EXPERIMENTAL AND RESULTS

### A. EXPERIMENTAL SETUP

To fairly compare the glioma grade prediction accuracy based on the proposed wavelet scattering features with that based on traditional features, experiments with different imaging modalities and different tumor regions were performed. Given all the samples (285), the training set and testing set were selected with the same ratio of HGG to LGG (3:1). That is, 57 samples were selected as testing set and 228 samples as training set. Since the number of HGG and LGG samples was not equal, sample balance was also considered by weighting the loss function of LR and SVM or the splitting criterion of RF with the class weight. Such sample equalization method was inspired by the work of King and Zeng [25] and implemented with Scikit-learn [26].

For the traditional radiomics, a total of 505 radiomic features were calculated for each patient with the package of Pyradiomics [27], including 9 shape features, 18 first-order features, and 478 textural features such as gray level co-occurrence matrix (GLCM) features, gray level size zone matrix (GLSZM) features, gray level run length matrix (GLRLM) features, neighboring gray tone difference matrix (NGTDM) features, gray level dependence matrix (GLDM) features and multiscale wavelet features. The detailed list of traditional features is given in Table S1 in the supplementary file. For the wavelet scattering-based radiomics, we obtained 505 feature maps ($S_0 = 1$, $S_1 = 6$, $S_2 = 498$) with scattering direction $L = 8$, wavelet decomposition scale $J = 6$ and scattering level $m = 2$. To enhance the contrast of the feature maps and keep the local invariance, a logarithmic operation followed by an averaging operation was performed on the wavelet scattering feature maps, which results in 505 wavelet scattering features in total for each patient. The wavelet scattering feature extraction was programed with Matlab 2016a and can be found in the work of [29]. To avoid redundancy and correlation of the large number of traditional or wavelet scattering-based features, PLS regression was used to reduce the feature dimension to 50 [30]. Specifically, the PLS regression parameters were learned using the training set, and then the feature dimension of the testing set was reduced with the well-learned PLS parameters. Finally, based on the selected features, three typical classifiers, namely, LR [31], SVM and RF [32], [33], were used to predict glioma grades.

Previous works showed that the performance of radiomic prediction is not only related to image features in the intratumoral region, but also to features in the peritumoral regions [34],[35]. Besides, multiple imaging modalities may provide more detailed information for promoting prediction accuracy [36]. To fairly evaluate the prediction performance of the proposed wavelet scattering-based radiomics, we performed the glioma grading on different tumor regions and using various imaging modalities, and quantitatively compared the proposed method with the traditional radiomics in terms of AUC and other metrics.

Regarding the dimension reduction of multimodal image features, the PLS model was performed firstly on the features extracted from each single image modality to reduce the number of features to 50. The dimension-reduced features are then combined (concatenated) and the combined features are finally reduced again with PLS to yield final 50 desired features.

### B. SUPERIORITY OF WAVELET SCATTERING FEATURES

To evaluate the superiority of wavelet scattering features, we tested the local invariance of wavelet scattering features by rotating and swapping a small block in the tumor region and adding Rayleigh noise. Then, we extracted traditional and wavelet scattering features from original and transformed tumor regions. The result is shown in Fig. 4.

We observe that rotating with 180° one small block (of size $11 \times 11$) in the tumor region results in an obvious change in traditional features, such as the cluster prominence (CP) feature in the first-order (FO) feature set (FO-CP), low gray level run emphasis (LGLRE) feature in the GLRLM feature set (GLRLM-LGLRE), and small area low gray level emphasis (SALGLE) in the GLSZM feature set (GLSZM-SALGLE). In contrast, in the case of WS features, when selecting one feature experiencing the biggest variation in rotation for illustration, we found that the WS features are very robust to the rotation: there is almost no difference between the WS features extracted from original image and those extracted from rotated image.

When we swapped two small blocks in the tumor region, traditional features, including autocorrelation (A) feature in the GLCM feature set (GLCM-A), large area high gray level emphasis (LAHGLE) feature in the GLSZM feature set (GLSZM-LAHGLE), and large dependence low gray level emphasis (LDLGLE) feature in the GLDM feature set (GLDM-LDLGLE), experience a big change. Such observation can also be found in the case after adding noise, which means that traditional features are very sensitive to noise and local variations. In contrast, swapping small blocks or adding noise has no influence on wavelet scattering features, as observed in the last row of Fig. 4.

To further quantitatively compare the changes of traditional features and wavelet scattering features before and after transformations, the sum of absolute changes in feature values was calculated. We estimated the difference in features by calculating the sum of absolute changes of features before and after transformation in the selected zone. For example, for a given pixel $i$, defining the value of a certain radiomic feature before transformation is $A_i$, and the value of which after transformation is $B_i$, then the change of such feature is defined as $\frac{1}{N}\sum_{i=1}^{N} abs(\frac{B_i - A_i}{A_i})$. Since the transformed feature value is related to the pre-transformation feature value, the normalization the feature value is not performed.

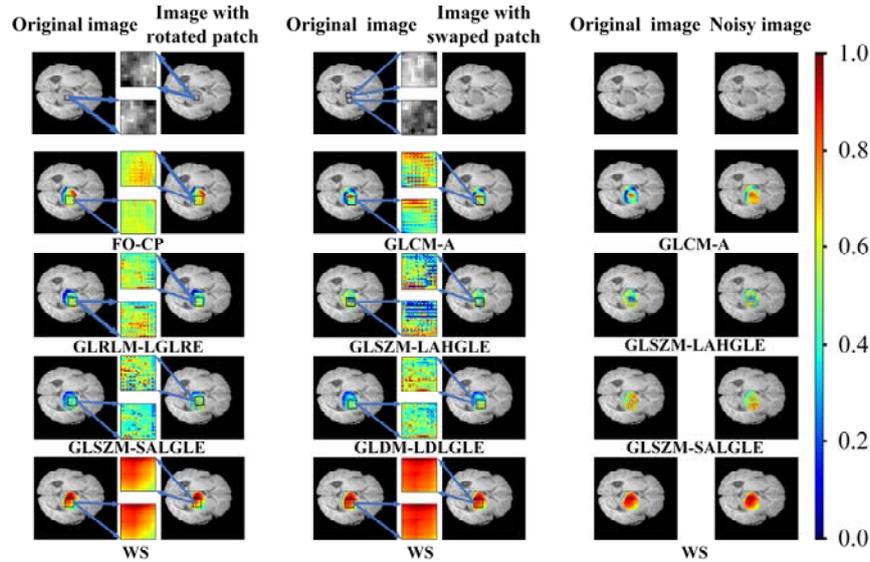

FIGURE 4. Superiority of wavelet scattering features in terms of local invariance and noise-robustness compared to traditional features.

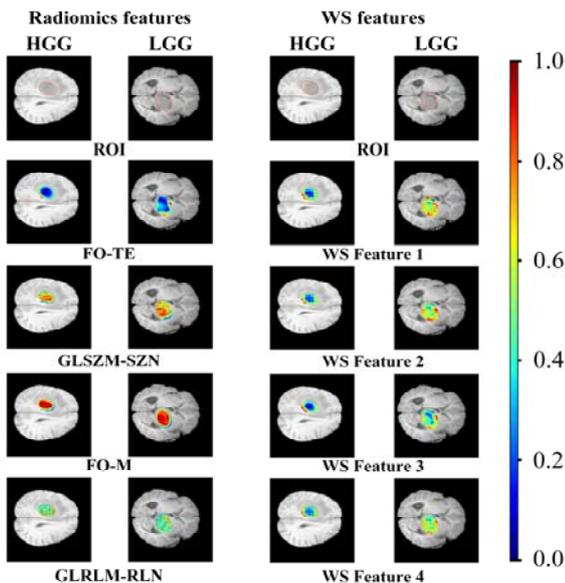

FIGURE 5. Features difference between LGG and HGG.

Rotating one block in tumor region leads to an absolute change of 7.72 in traditional features and 0.03 in wavelet scattering features. Swapping two blocks in tumor region generates a change of 4.9 in traditional features and 0.04 in wavelet scattering features. Adding noise results in a change of 42.27 in traditional features and 1.35 in wavelet scattering features.

All these results demonstrate that wavelet scattering features outperforms traditional features in terms of local invariance and noise robustness, which is very important for promoting the classification accuracy.

In addition, we demonstrated further the traditional radiomic features and WS features for LGG and HGG in Fig. 5. We observe that WS features for LGG and HGG are totally different, however, the difference in radiomic features between LGG and HGG are not obvious. This illustrates that WS features will be beneficial for classifying the LGG and HGG.

## C. QUANTITATIVE COMPARISON OF GLIOMA GRADING ON DIFFERENT REGIONS

As illustrated in Fig. 1, the labeled ROIs include intratumoral and peritumoral regions. We extracted traditional features and wavelet scattering features first from intratumoral region and then from both inratumoral and peritumoral regions. To avoid the influence of imaging modality, in this experiment, the single modal images, namely, T1, T1-CE, T2 and Flair were considered separately. The ROCs of the three classifiers with different features extracted from different regions in T1 images are shown in Fig. 6, and the ROCs curves for other modalities are given in Fig. S1- Fig. S3 in the supplementary file.

In these ROC curves, the middle black curve indicates the dividing line with an AUC of 0.5, the green curve the ROC obtained with traditional features, and the red curve the ROC obtained with wavelet scattering features. It can be easily observed that the glioma grading accuracy with wavelet scattering features is much better than that with traditional features, especially for SVM and LR classifiers. In addition, comparing Fig. 6(a) and 6(b) shows that the image features extracted from the peritumoral regions are helpful to promote the glioma grading accuracy.

To further quantitatively compare the prediction performance of glioma grading with different features, the accuracy, sensitivity, specificity and AUC are given in Table II for T1 modality and in Table S2-Table S4 for other single modalities. Clearly, the metrics have much higher values using wavelet scattering features than using traditional features. In the prediction with intratumoral features, compared to the wavelet-based method, the AUC of wavelet scattering-based prediction is increased by about 12.9%, 11.7% and 8.4% for SVM, LR and RF, respectively. As to the prediction with both intratumoral and peritumoral image features, the AUC obtained with wavelet scattering-based features is increased by about 7.8%, 12.9% and 9.1 % for SVM, LR and RF, respectively. We observe that the features extracted from peritumoral regions promote the prediction accuracy, especially for the traditional radiomics. In other words, the peritumoral features decreased the difference in prediction accuracy between traditional features-based and wavelet scattering features-based methods.

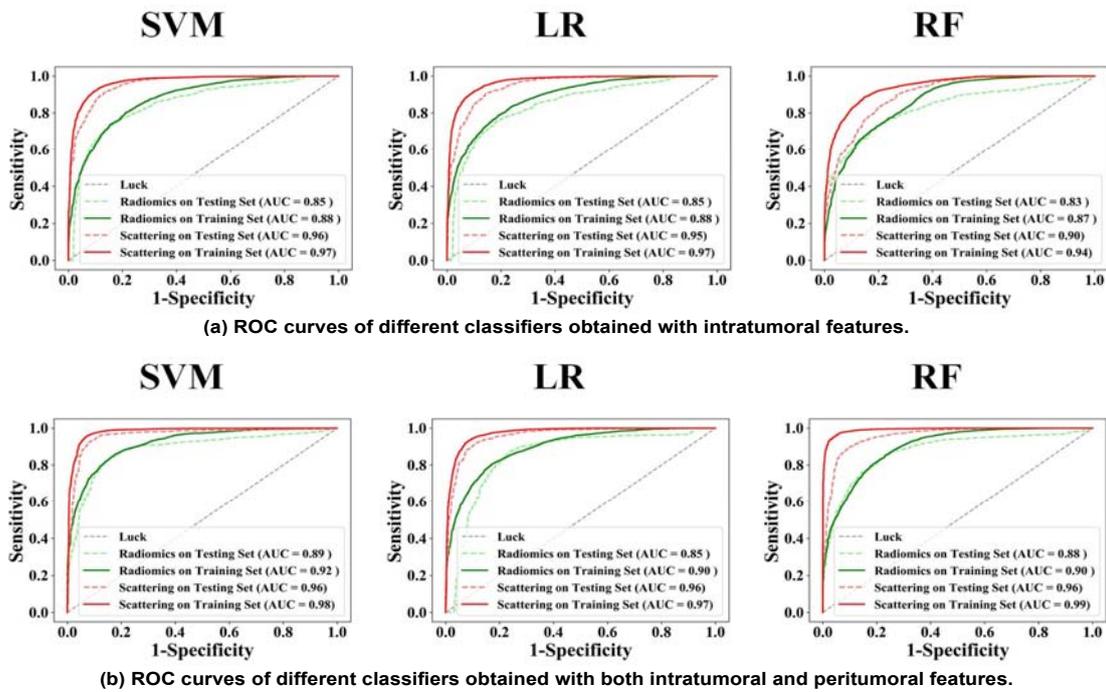

(a) ROC curves of different classifiers obtained with intratumoral features.

(b) ROC curves of different classifiers obtained with both intratumoral and peritumoral features.

**FIGURE 6.** ROC curves of different classifiers obtained with traditional features and wavelet scattering features extracted from different ROIs of T1 images.

TABLE II

QUANTITATIVE COMPARISON OF GLIOMA GRADINGS IN THE CASE OF USING DIFFERENT FEATURES EXTRACTED FROM DIFFERENT REGIONS IN T1 IMAGES.

| ROI | Features | Model | Accuracy | Sensitivity | Specificity | AUC |
|---|---|---|---|---|---|---|
| Intratumoral | Traditional | SVM | 0.79 | 0.81 | 0.73 | 0.85 |
| | | LR | 0.78 | 0.79 | 0.73 | 0.85 |
| | | RF | 0.78 | 0.84 | 0.62 | 0.83 |
| | Wavelet scattering | SVM | 0.90 | 0.94 | 0.81 | 0.96 |
| | | LR | 0.89 | 0.92 | 0.80 | 0.95 |
| | | RF | 0.86 | 0.93 | 0.65 | 0.90 |
| Intratumoral + Peritumoral | Traditional | SVM | 0.84 | 0.86 | 0.81 | 0.89 |
| | | LR | 0.82 | 0.84 | 0.78 | 0.85 |
| | | RF | 0.83 | 0.84 | 0.78 | 0.88 |
| | Wavelet scattering | SVM | 0.92 | 0.92 | 0.91 | 0.96 |
| | | LR | 0.90 | 0.90 | 0.89 | 0.96 |
| | | RF | 0.91 | 0.94 | 0.82 | 0.96 |

*D. QUANTITATIVE COMPARISON OF GLIOMA GRADING WITH DIFFERENT MODALITIES*

To investigate the glioma grading performance of traditional features-based and wavelet scattering features-based methods with multimodal images, we first extracted the image features from different combinations of multimodal images, including the combination of arbitrary two modalities, three modalities and four modalities. Then, glioma grading based on these features was performed using different classifiers. The ROC curves for the classification of glioma grades, which were obtained with traditional features and wavelet scattering features extracted from different ROIs of multimodal images（combination of T1, T1-CE, and T2 images), are illustrated in Fig. 7. The ROC curves for the combination of the other modalities were given in Fig. S4 - Fig. S13. We can see that the grading accuracy based on the wavelet scattering features extracted from the multimodal images is even higher than that based on the traditional features, whereas this superiority is not as evident as in the grading with single-modality images.

The corresponding quantitative results are given in Table III, and the metrics for the other combinations of multimodalities are illustrated in the supplementary results (from Tables S5 to S14). The features extracted from multimodal images indeed increase the glioma grading accuracy, AUC and sensitivity, but the AUC difference between traditional features and wavelet scattering features is not so obvious. Nevertheless, the specificity and sensitivity obtained with wavelet scattering features are much higher than those obtained with traditional

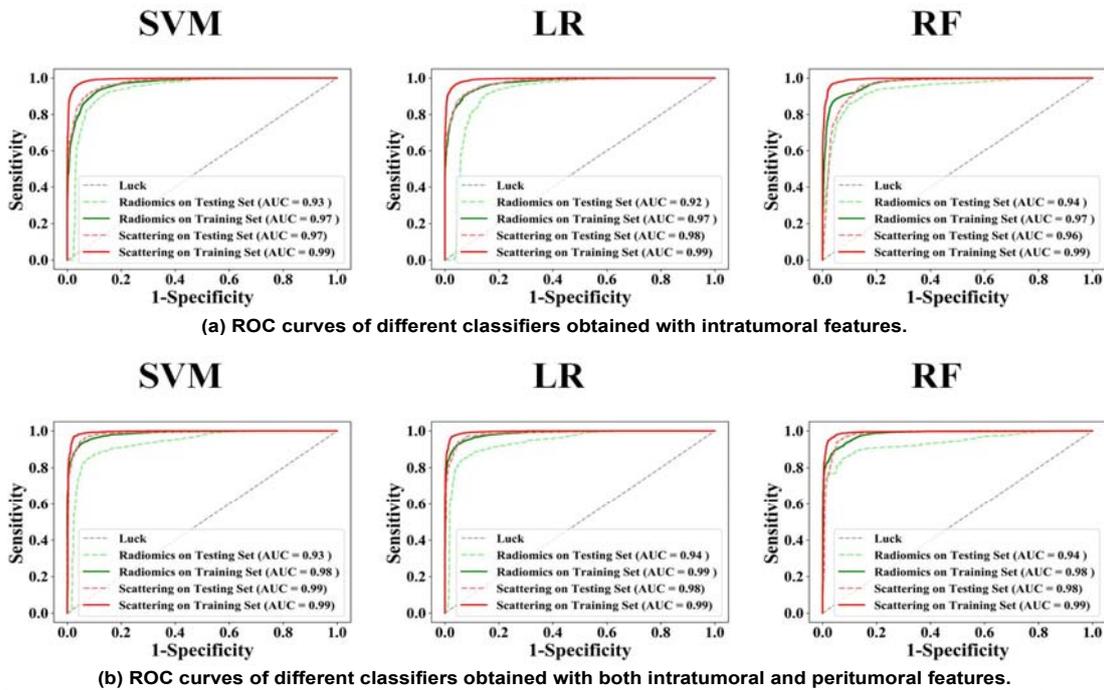

(a) ROC curves of different classifiers obtained with intratumoral features.

(b) ROC curves of different classifiers obtained with both intratumoral and peritumoral features.

**FIGURE 7.** ROC curves of different classifiers obtained with traditional features and wavelet scattering features extracted from different ROIs of multimodal images.

TABLE III

QUANTITATIVE COMPARISON OF GLIOMA GRADINGS IN THE CASE OF USING DIFFERENT FEATURES EXTRACTED FROM BOTH INTRATUMORAL AND PERITUMORAL REGIONS OF MULTIMODAL IMAGES.

| ROI | Features | Model | Accuracy | Sensitivity | Specificity | AUC |
|---|---|---|---|---|---|---|
| Intratumoral | Traditional | SVM | 0.88 | 0.89 | 0.85 | 0.93 |
| | | LR | 0.86 | 0.85 | 0.87 | 0.92 |
| | | RF | 0.87 | 0.87 | 0.87 | 0.94 |
| | Wavelet scattering | SVM | 0.92 | 0.94 | 0.87 | 0.97 |
| | | LR | 0.92 | 0.94 | 0.87 | 0.98 |
| | | RF | 0.93 | 0.95 | 0.86 | 0.96 |
| Intratumoral + Peritumoral | Traditional | SVM | 0.87 | 0.85 | 0.91 | 0.93 |
| | | LR | 0.87 | 0.86 | 0.92 | 0.94 |
| | | RF | 0.88 | 0.88 | 0.88 | 0.94 |
| | Wavelet scattering | SVM | 0.94 | 0.94 | 0.95 | 0.99 |
| | | LR | 0.94 | 0.93 | 0.95 | 0.98 |
| | | RF | 0.94 | 0.94 | 0.93 | 0.98 |

features, which demonstrates that the proposed method decreases simultaneously false positive and false negative errors in glioma grading.

### E. COMPARISON WITH DEEP LEARNING-BASED METHODS

Besides the comparison with the traditional radiomic features, to further validate the superiority of our methods, we compared our methods with the radiomics performed by others and some deep learning-based models in terms of accuracy, sensitivity, specificity, and AUC. For the fair comparison, we use the same dataset that comes from MICCAI BraTS 2017.

Table IV shows comparison results. It is clearly observed that, comparing the radiomics, the accuracy, sensitivity, specificity, and accuracy are all increased a lot. With respect to the deep learning-based method, the prediction accuracy is also increased. This verified that the transformation-invariant property of WS features can help to promote prediction accuracy of the glioma grading.

TABLE IV
STATE-OF-THE-ART METHODS VS THE PROPOSED SCHEME.

| | Methods | Accuracy | Sensitivity | Specificity | AUC |
|---|---|---|---|---|---|
| Radiomics | Cho et al. [16] | 0.89 | 0.96 | 0.68 | 0.90 |
| | Decuyper et al. [39] | 0.90 | 0.90 | 0.89 | 0.96 |
| Deep learning-based methods | Ge et al. [40] | 0.89 | - | - | - |
| | Ge et al. [41] | 0.91 | - | - | - |
| | Ali et al [42] | 0.92 | - | - | - |
| Ours | Table III | **0.94** | **0.94** | **0.95** | **0.99** |

## IV. DISCUSSION

The proposed glioma grade prediction method is based on local invariant features extracted from wavelet scattering network instead of traditional features. The experiments on different tumor regions with different imaging modalities showed that the proposed method reaches an AUC of 0.90 at least, which demonstrates its effectiveness and superiority.

Traditional radiomic methods used handcrafted features to train classification models. Such features include shapes, textural, wavelet and statistical information, which are susceptible to image intensity variation and image deformation. As a result, the prediction accuracy based on these features is influenced. To cope with such problem and in view of the interest of wavelet scattering transform for the extraction of invariant image features, we replaced traditional radiomic features by wavelet-scattering radiomic features to predict glioma grades. The experimental results showed that the use of those invariant features allows us to better represent image properties and thus distinguish more effectively glioma grades. This can be reflected in the samples clustering results, as shown in Fig. 8, which was obtained using the Consensus Cluster Plus package in bioinformatics analysis software under R language [37],[38].

The consensus matrix $M$ is usually used to reflect the quality and robustness of clustering. The horizontal or vertical axis of $M$ represents the sample index. If we have $N$ samples, the size of $M$ is $N \times N$. The value of each matrix element $M_{ij}$ is between 0 and 1, representing the probability that the sample $i$ and sample $j$ are clustered into the same class during multiple clustering tests. In our case, the blue color indicates the probability value of 1 and the white color the probability value of 0. The cleaner the consensus matrix, the better the clustering results. Considering the distribution of datasets used in this work, with 75 patients presenting LGG and 210 patients presenting HGG, the ratio of positive (high-grade) to negative (low-grade) samples is about 3:1, therefore the ratio of the corresponding clusters should be also about 3:1.

When using single modality T1-weighted images (Fig. 8(a)), the ratio of cluster 1 to cluster 2 grouped with wavelet scattering features is about 3:1, whereas the ratio obtained with traditional features is about 3:2. In the case of intratumor features, the clustering accuracy of traditional radiomics for LGG and HGG is respectively 50.2% and 50.5%, and that of wavelet scattering-based radiomics for LGG and HGG is respectively 50.0% and 67.7%. In the case of both intra- and peri-tumoral features, the clustering accuracy of traditional radiomics for LGG and HGG is respectively 45.0% and 67.1%, and that of wavelet scattering radiomics for LGG and HGG is respectively 67.5% and 71.1%. All that clearly explains why wavelet scattering features are superior to traditional features.

In the case of multimodal image features (Fig. 8(b)), when using traditional features, prediction accuracy increases. This is reflected by both the ratio of two clusters (closer to 3:1 knowing that the ratio of HGG to LGG is 3:1) and the clustering accuracy. Indeed, with intratumoral features, the clustering accuracy for LGG and HGG is respectively 70.0% and 72.7%. With both intra- and peri-tumoral features, the clustering accuracy for LGG and HGG is respectively 75.1% and 80.8%. Always in the case of multimodal image features, when using wavelet scattering features, the clustering consensus maps with multimodal image features become much clearer, and the clustering accuracy for LGG and HGG passes respectively from 75.1% to 90.02% and from 80.8% to 95.6%, when using both intra- and peri-tumoral features.

All these results explain why the features extracted from wavelet scattering have correctly represented data distribution and are more appropriate for glioma grading, and why the prediction accuracy based on wavelet scattering features is higher than that based on traditional features. They also explain why taking into account peritumoral regions makes consensus matrix maps much cleaner and consequently prediction more accurate.

As to the influence of imaging modalities, we observed that the prediction with T1-enhanced mages was the best if considering only single modality. The combination of T1-enhanced and T2-weighted images was the best if two modalities were used, and the combination of T1-enhanced, T2- and T1-weighted images was the best if three modalities were required. Comparing Tables II and III show that the number of modalities used in the predictions does not influence greatly the wavelet scattering-based radiomics but impacts more the traditional radiomics, which implies that with the wavelet scattering-based radiomics, we are able to grade glioma accurately just with one single modality. This is beneficial for reducing acquisition time.

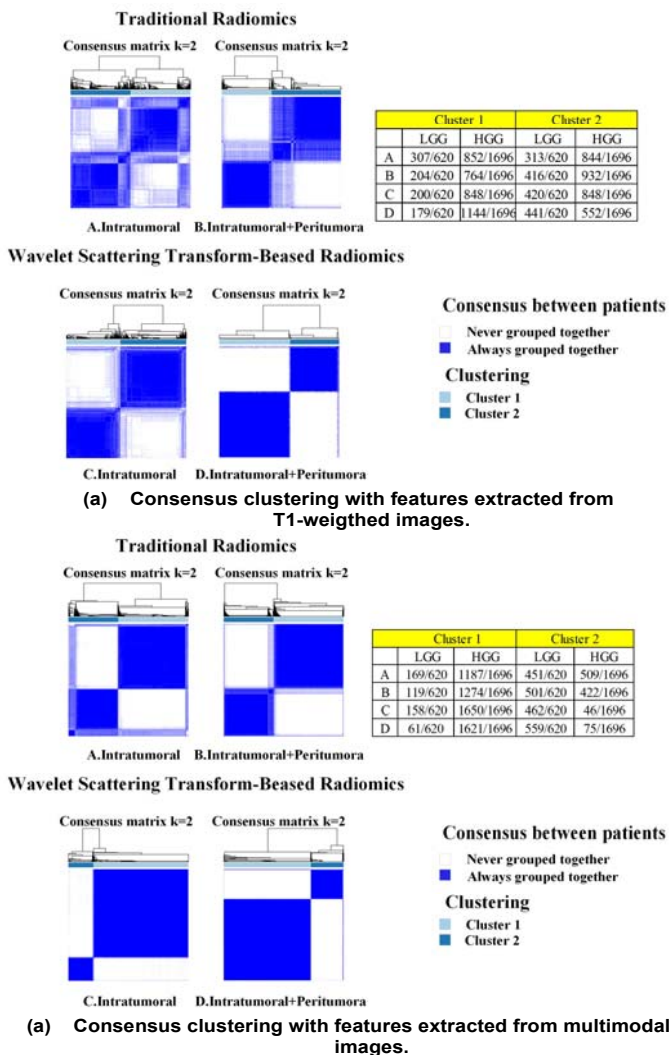

**(a)** Consensus clustering with features extracted from T1-weigthed images.

**(a)** Consensus clustering with features extracted from multimodal images.

 Consensus clustering for traditional features and wavelet scattering features extracted from different ROIs in both single modal and multiple modal images.

Although the AUC, sensitivity and specificity obtained with the proposed method are greatly increased, there are still several limitations in the present work. Firstly, the dataset used is a public open source dataset, and the training cohort and testing cohort come from the same group. In the future, it would be interesting to test with different cohorts to further evaluate the proposed method. Moreover, the radiomic prediction being performed on segmented ROIs, the quality of segmentation can influence the subsequent prediction. To deal with this issue and in light of the promising deep learning models [45], we may combine deep learning models and wavelet scattering to achieve glioma grading without the requirement for image segmentation.

## V. CONCLUSIONS

We have proposed a novel wavelet scattering-based radiomic method to predict noninvasively and accurately glioma grades before surgery. The method is based on the use of local invariant features extracted from wavelet scattering transform instead of traditional features as used in existing radiomic methods. The results showed that the high-dimensional image features extracted from wavelet scattering-based radiomics improve greatly the accuracy of glioma grading. Furthermore, peritumoral features are beneficial for glioma grading. All that suggests the potential use of the proposed method for computer-aided glioma diagnosis.